# Adversarial Machine Learning Attacks on Condition-Based Maintenance Capabilities

Hamidreza Habibollahi Najaf Abadi

Center for Advanced Life Cycle Engineering (CALCE)
University of Maryland
College Park, MD, 20742 USA

**Abstract**

Condition-based maintenance (CBM) strategies exploit machine learning models to assess the health status of systems based on the collected data from the physical environment, while machine learning models are vulnerable to adversarial attacks. A malicious adversary can manipulate the collected data to deceive the machine learning model and affect the CBM system's performance. Adversarial machine learning techniques introduced in the computer vision domain can be used to make stealthy attacks on CBM systems by adding perturbation to data to confuse trained models. The stealthy nature causes difficulty and delay in detection of the attacks. In this paper, adversarial machine learning in the domain of CBM is introduced. A case study shows how adversarial machine learning can be used to attack CBM capabilities. Adversarial samples are crafted using the Fast Gradient Sign method, and the performance of a CBM system under attack is investigated. The obtained results reveal that CBM systems are vulnerable to adversarial machine learning attacks and defense strategies need to be considered.

## 1. Introduction

Maintenance strategies can be classified into corrective maintenance and preventive maintenance. Corrective maintenance is a strategy that is used to repair (or replace) the component after it fails, while preventive maintenance is carried out before break downs. Condition-based maintenance (CBM), a type of preventive maintenance, recommends maintenance decisions based on the information collected through condition monitoring process [1]. Condition monitoring equipment includes sensor nodes that measure different parameters, such as pressure, temperature, speed, vibration, etc. These measurements are processed in the decision layer of the CBM system after passing through a network in order to assess the health status of a system (Figure 1).

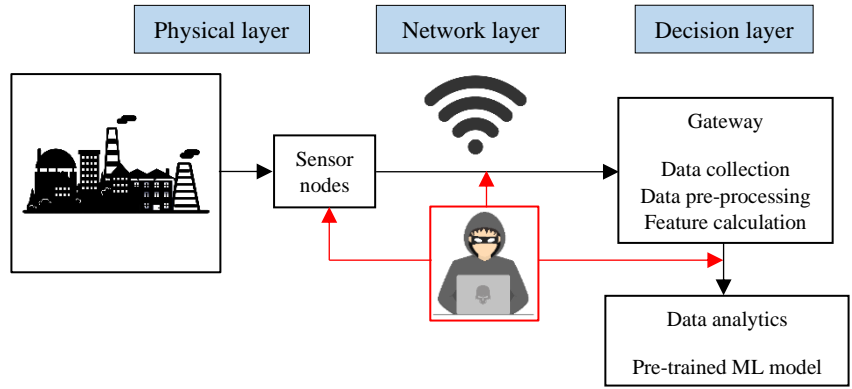

Figure 1. Condition-based maintenance architecture and threat model.

Machine learning (ML) techniques are widely used to perform CBM strategies [2]. For example, the diagnostic capability of a CBM system that takes advantage of ML techniques is a pattern recognition algorithm that maps the information in the measurement space or features in the feature space to machine faults in the fault space and is supposed to identify the health status of a machine [2]. Although exploiting ML in CBM systems has brought efficiency in automated detection of machines' health status, the ML models are vulnerable to adversarial attacks, and industries have not considered the strategies to protect their ML systems [3]. Misleading CBM systems could lead to delay in detecting imminent failures that may result in financial loss, and even loss of life, as CBM systems are being used in many critical industries.

The act of deploying attacks towards ML-based systems is known as Adversarial Machine Learning (AML) [4]. Such adversarial attacks have been investigated in the field of computer vision, where adding slight perturbations – insignificant to human eye – to an image can cause a classifier model to misclassify it (Figure 2) [5]. The idea behind the AML is that the imperfections in the training phase of ML models make them powerless to adversarial samples.

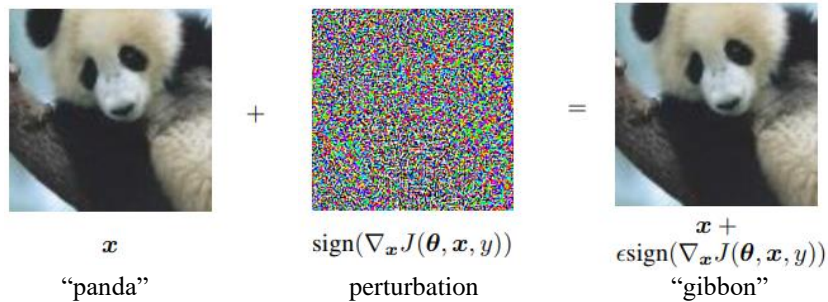

Figure 2. Crafting adversarial examples in the computer vision domain *[6]*.

In the case of CBM systems, AML can be used to manipulate the data obtained from sensors by adding perturbations in order to cause data related to faulty conditions to be identified as healthy and vice versa. AML intends to exploit the fact that the trained ML models have "blind spots" among the data points that they have seen before in the training phase. So, adding perturbations to the unseen data points can cause the models to cross a decision boundary and classify the data as incorrect classes [4]. Usually, the perturbations to deceive an ML model are designed to be small in magnitude so that a human being may not be able to understand that anything has changed [7]. The manipulated data can be injected – fault injection attack –at each of the CBM layers (Figure *1*). Deceiving the ML algorithm affects the integrity of the model and a high rate of errors can cause the model to become unusable.

Considering the application of CBM capabilities in critical systems, they have become an attractive target to attackers. Also, as ML-based CBM systems become more widely deployed, the adversary motivation for defeating them increases. Therefore, CBM systems need to be evaluated against AML attacks. This paper investigates generating stealthy attacks that evade classification algorithms used for diagnostic purposes in a CBM system. This study shows how AML can be used to attack CBM capabilities. Section 2 of this paper defines the specifications of AML attacks. The experimental results showing the impact of AML attacks on a CBM system are presented in Section 3. Section 4 concludes the paper.

2. **Adversarial Machine Learning Attacks**

Two main types of AML attacks are evasion attacks and poisoning attacks. Evasion attacks can be known as attacks on the decisions made by trained models by making changes to the collected data from the environment. An evasion attack introduces perturbation to input to cause its misclassification [8]. Unlike the evasion attacks, which target the trained models during the testing phase, poisoning attacks occur during the learning phase by modifying a part of the data used for training to cause the learning algorithm to make poor choices [9].

This paper focuses on evasion attacks to CBM systems. Given a trained ML model *f*, adversaries add a perturbation $\varepsilon$ to sample $x$ to generate adversarial sample $x^*$ such that $C_{x+\varepsilon} \neq C_x$, where C is the class predicted by the victim model (Equation 1). To cause delay in detecting the attacks, the perturbation should be slight. The difference between crafted adversarial sample and the original sample can be evaluated by a distance metric $\|x^* - x\|$. Frequently, this distance metric is the $L_\infty$ norm, which measures the maximum allowable perturbation on any feature; the $L_0$ norm, which determines the maximum number of features that can be changed; or finally the $L_2$ norm, which is the Euclidian distance between $x$ and $x^*$ [8].

$$\min_{x'} \|x^* - x\|$$
$$\text{s.t. } f(x^*) = C_{x+\varepsilon}$$
$$f(x) = C_x$$
$$C_{x+\varepsilon} \neq C_x$$
(1)

An AML attack can be a targeted attack. It means that the goal of the attack is to change the collected data in a way that they will be classified as a specific target class. The targeted attacks happen on supervised learning, which attackers aspire to induce classification according to a target class function [9]. On the other hand, an untargeted attack does not target a specific target class; however, it attempts to maximize errors to affect the model's reliability.

Based on the knowledge an adversary may have, attacks can be categorized as white-box, black-box, and grey-box attacks [4]. In white-box attacks, the adversary has access to the training dataset and has complete knowledge about the ML model, including the architecture, model parameters, and features. However, the assumption that the attacker has complete information about the system is inaccurate in the real world [9]. In contrast, black-box attacks assume the adversary has no information about the ML model. In black-box attacks, the adversary does not have access to the same training data as the ML model. Still, it is usually assumed that they can intercept the system communication and collect a dataset similar to the training dataset or set an identical environment to generate a similar dataset [10].

Black-box attacks rely on the transferability of adversarial samples, which means that adversarial examples that affect one model often affect another model, even if the two models are different or were trained on separate but similar training datasets [11]. In other words, the adversarial samples generated by a local learning model could confuse other models [10]. Therefore, an attacker can train its substitute model using a similar dataset, craft adversarial samples against the substitute model, and transfer them to a victim model in order to impact its performance, without knowing anything about the victim model.

## 3. Condition-Based Maintenance System Case Study: Motor Bearings

In this study, a bearing was considered as the component being maintained by a CBM system. The bearing dataset is provided by the Case Western Reserve University [12]. The experimental setup schematic for collecting the data is shown in Figure 3. There are a two hp induction motor (left), a torque transducer (middle), and a dynamometer (right). Single point faults were made in the bearings' inner raceway, rolling elements, and outer raceway. The faults have diameters of 7 mils, 14 mils, 21 mils, 28 mils, and 40 mils. Vibration data were collected for motor loads from 0 to 3 hp and motor speed around 1,750 rpm using two accelerometers at the drive end and fan end of the motor. Also, two sampling frequencies of 12 kHz and 48 kHz were used. The vibration data collected by the drive end accelerometer with 12 kHz was used in this analysis.

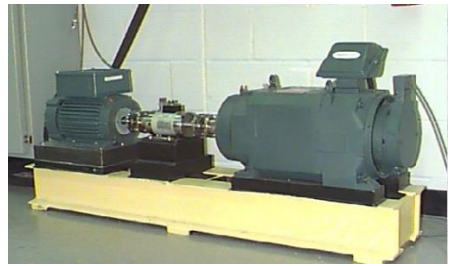

Figure 3. Experimental setup for collecting the bearing dataset *[12]*.

### 3.1. Methodology

Twelve features, including Clearance factor, Crest factor, Impulse factor, Kurtosis, mean, peak, Root Mean Square (RMS), Shape factor, Skewness, and standard deviation from time-domain plus peak amplitude and peak frequency from frequency-domain, were extracted for 0.1-s time windows from the bearing data set to generate training and test datasets. Descriptions of the features can be found in the literature [22]. Details about the size of the data are summarized in Table 1.

Table 1. The size of the dataset.

| Label | Sample Size |
|---|---|
| Ball | 8162 |
| Inner race | 8162 |
| Outer race | 2056 |
| Normal | 3371 |

According to the "no free lunch" theorem there is no best learning algorithm for all possible problems [13] and an appropriate algorithm for a specific task should be selected based on its performance on that problem. Therefore, nine different classification algorithms in the Scikit-learn ML library, including random forest, multilayer perceptron, decision tree, k-nearest neighbors, non-linear support vector machine, linear support vector machine, quadratic discriminant analysis, naïve Bayes, and AdaBoost, were evaluated without adversarial attacks. Then, adversarial attacks were applied to investigate the impact on the models with f1 scores higher than 0.95.

Grey-box evasion attacks were considered in the analysis, assuming the attacker has knowledge about the used features by the CBM model and has a training dataset similar to the training dataset that was used to train the CBM model but does not know about the used algorithm in the CBM system. The attacker trains its substitute model using the available dataset and craft adversarial samples relying on ML algorithms' transferability. The crafted sample was then injected into the system at the decision layer (Figure 1) to deceive the system and affect its performance. In this paper, the substitute algorithm/model means the algorithm/model used by a hypothetical adversary to craft adversarial samples, and the victim algorithm/model is related to the CBM system under an evasion attack.

Using 40% of the dataset, a deep neural network with three hidden layers with the ReLU activation function consist of 64, 64, and 32 neurons and an output layer with the Softmax activation function was trained as the substitute model. The adversarial samples were then crafted by adding a range of perturbations (from 0.01 to 0.05) to the 20% of the data. The remaining 40% of the dataset was used to evaluate (k-fold cross-validation) and train the CBM victim models. The f1 score for the different victim models in classification of the adversarial samples was measured to investigate CBM systems' vulnerability to AML attacks.

There are different methods to generate adversarial samples. Such approaches vary in complexity and their performance, such as the Fast Gradient Sign Method (FGSM), Jacobian-Based Saliency Map Attack, Basic Iterative Method, and Carlini Wagner. In this study, FGSM in the Adversarial Robustness Toolbox [14] was used for crafting adversarial samples.

FGSM was proposed by Goodfellow et al. [6] to craft adversarial samples of images to deceive the GoogLeNet model, a convolutional neural network. The FGSM modifies the data by adding a specific amount of perturbation noise obtained using the gradient of the cost function (Equation 2). The FGSM is a one-shot method as the adversarial samples are generated through a single step of the computation.

$$x^* = x + \varepsilon \, sign \, (\nabla_x J(\theta, x, y)) \qquad (2)$$

Where $\theta$ is the model parameters, $x$ indicates the input data, $y$ denotes the labels, $\varepsilon$ is the hyper-parameter that determines the extent of the perturbation, and $J$ is the cost function.

### 3.2. Result

Using the k-fold cross-validation (k=5), different algorithms' performance in identifying the bearing health status was calculated (Table 2). F1 score for all the algorithms, except AdaBoost, was higher than 0.9. Random forest, multilayer perceptron (consisting of 1 hidden layer with 32 neutrons), decision tree, k-nearest neighbors, and kernel support vector machine algorithms showed a F1 score of more than 0.99. Quadratic discriminant analysis obtained an score of 0.96, and the linear support vector machine got an score of 0.95.

Table 2. Classification F1 score for different algorithms using 5-fold cross-validation

| Algorithm | F1 score |
|---|---|
| Random Forest | 1 |
| Multilayer Perceptron (3 Layers) | 1 |
| Decision Tree | 1 |
| K-Nearest Neighbors | 0.99 |
| Support Vector Machine (Kernel: RBF) | 0.99 |
| Quadratic Discriminant Analysis | 0.96 |
| Support Vector Machine (Linear) | 0.95 |
| Naïve Bayes | 0.91 |
| AdaBoost | 0.57 |

It is assumed the adversary does not have any information about the CBM algorithm and the used ML model but knows about the features and has access to a set of training data similar to the data used to train the CBM algorithm. A deep neural network was used as the substitute model to craft adversarial samples by using FGSM to investigate the top 7 models (models with F1 scores higher than 0.95) under evasion attacks. Using a substitute algorithm different from the victim algorithms allows the evaluation of adversarial attacks' transferability by applying the adversarial samples created with a model on the other models.

### 3.2.1. Fast Gradient Sign Method

FGSM is a method to generate adversarial samples for evasion attacks on trained models. In FGSM, the goal is to create samples that maximize the loss value. To do that, the gradient of loss function is taken for each input data point to find the contribution level of each feature in the loss value and add a perturbation accordingly. For a successful attack, an appropriate amount of perturbation needs to be selected. The lower extent of perturbations will lead to stealthier adversarial attacks that are more difficult to detect. Table 3 shows an example of the perturbed features for an adversarial sample.

Table 3. An example of how features are perturbed by using FGSM ($\varepsilon$: extent of the perturbation).

| Feature | Original data point (Class: Inner Race) | Perturbed point ($\varepsilon=0.03$) (Predicted class by Random Forest: Outer Race) |
|---|---|---|
| Clearance factor | 0.18 | 0.21 |
| Crest factor | 0.43 | 0.40 |
| Impulse factor | 0.23 | 0.26 |
| Kurtosis | -0.08 | -0.05 |
| Mean | -0.10 | -0.13 |
| Peak | -0.59 | -0.62 |
| RMS | -0.65 | -0.68 |
| Shape factor | 0.17 | 0.20 |
| Skewness | 1.46 | 1.43 |
| Standard deviation | -0.65 | -0.68 |
| Peak frequency amplitude | -0.63 | -0.60 |
| Peak frequency | 1.01 | 0.98 |

Except the k-nearest neighbors model, all the models experienced a performance drop in the classification of the crafted adversarial samples, although the drops' amount was different (Figure *4*). The quadratic discriminant analysis showed the most significant performance drop among all models, which shows this model's weakness against adversarial attacks. The random forest that previously showed a high F1 score (1) in identifying the system's health status obtained a F1 score of 0.83 after adding a slight perturbation with the extent of 0.05. In the same situation, multilayer perceptron reached a F1 score of 0.95, and the decision tree got a F1 score of 0.91, while for the original dataset, they obtained a F1 score equal to 1. The k-nearest neighbors model was the most robust algorithm. Also, the kernel support vector machine showed robustness relatively with just 0.01 drop in the score.

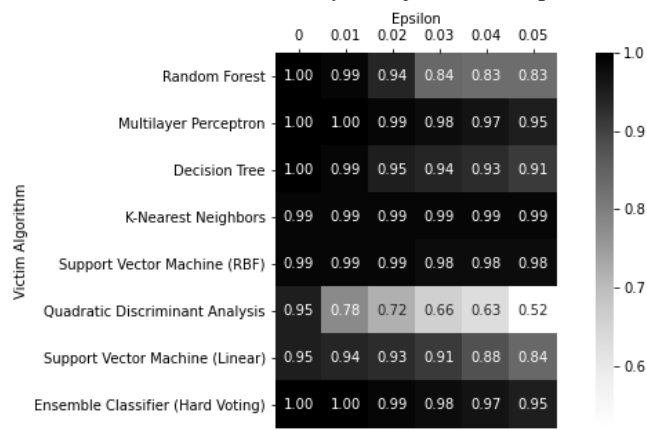

Figure 4. F1 score for different machine learning algorithms on adversarial samples crafted using FGSM ($\varepsilon$: extent of the perturbation).

Not all the algorithms misclassified the same adversarial data points. For example, the random forest algorithm misclassified a high number of adversarial samples related to the normal class as ball class (Figure 5a). In contrast, multilayer perceptron and decision tree algorithms were able to identify all adversarial samples associated with the normal class correctly (Figure 5a and b). On the other hand, the decision tree and the multilayer perceptron models identified ball class data points more accurately than the decision tree model.

|  |  | Predicted Class | | | | |
|---|---|---|---|---|---|---|
|  |  | Ball | Inner Race | Outer Race | Normal |  |
| Actual Class | Ball | 1634 | 2 | 0 | 3 | 99.69% |
|  | Inner Race | 3 | 1463 | 178 | 0 | 88.99% |
|  | Outer Race | 5 | 16 | 389 | 0 | 94.88% |
|  | Normal | 498 | 0 | 0 | 160 | 24.32% |
|  |  | 76.35% | 98.78% | 68.61% | 98.16% |  |

(a)

|  |  | Predicted Class | | | | |
|---|---|---|---|---|---|---|
|  |  | Ball | Inner Race | Outer Race | Normal |  |
| Actual Class | Ball | 1633 | 2 | 0 | 4 | 99.63% |
|  | Inner Race | 2 | 1626 | 16 | 0 | 98.90% |
|  | Outer Race | 5 | 30 | 375 | 0 | 91.46% |
|  | Normal | 0 | 0 | 0 | 658 | 100% |
|  |  | 99.57% | 98.07% | 95.91% | 99.39% |  |

(b)

|  |  | Predicted Class | | | | |
|---|---|---|---|---|---|---|
|  |  | Ball | Inner Race | Outer Race | Normal |  |
| Actual Class | Ball | 1603 | 20 | 15 | 1 | 97.80% |
|  | Inner Race | 18 | 1439 | 179 | 0 | 87.95% |
|  | Outer Race | 4 | 3 | 397 | 0 | 98.27% |
|  | Normal | 0 | 0 | 0 | 672 | 100% |
|  |  | 98.65% | 98.43% | 67.17% | 99.85% |  |

(c)

Figure 5. Confusion matrix for a) random forest, b) multilayer perceptron, and c) decision tree classification models on adversarial samples crafted using FGSM ($\varepsilon$: 0.03).

### 3.3. Discussion

The FGSM was able to craft adversarial samples to mislead the CBM system. The misclassification of adversarial data points by the trained models can be observed by visualizing the data using t-distributed stochastic neighbor embedding (t-SNE) dimension reduction. In Figure 6a, data points related to the normal class, a portion of the ball data points, and a portion of the inner race data points have made clusters with distinct boundaries. While parts of the ball data points and inner race data points as well as the data points related to the outer race class have overlapped. Although the high F1 score obtained in the classification of the original data points (1) represents that there are no real overlaps of different classes in the feature space and the observed overlaps in 2D visualization are resulted from the data points that were located close to the boundary decisions.

The idea behind crafting adversarial samples is to perturb the data points close to decision boundaries to make them cross the boundaries and cause misclassifications without creating considerable changes in the data. This matter can be seen in 2D visualization of the adversarial data points crafted using FGSM (Figure 6b) where the overlap areas consist of misclassified data points, while other clusters are classified correctly.

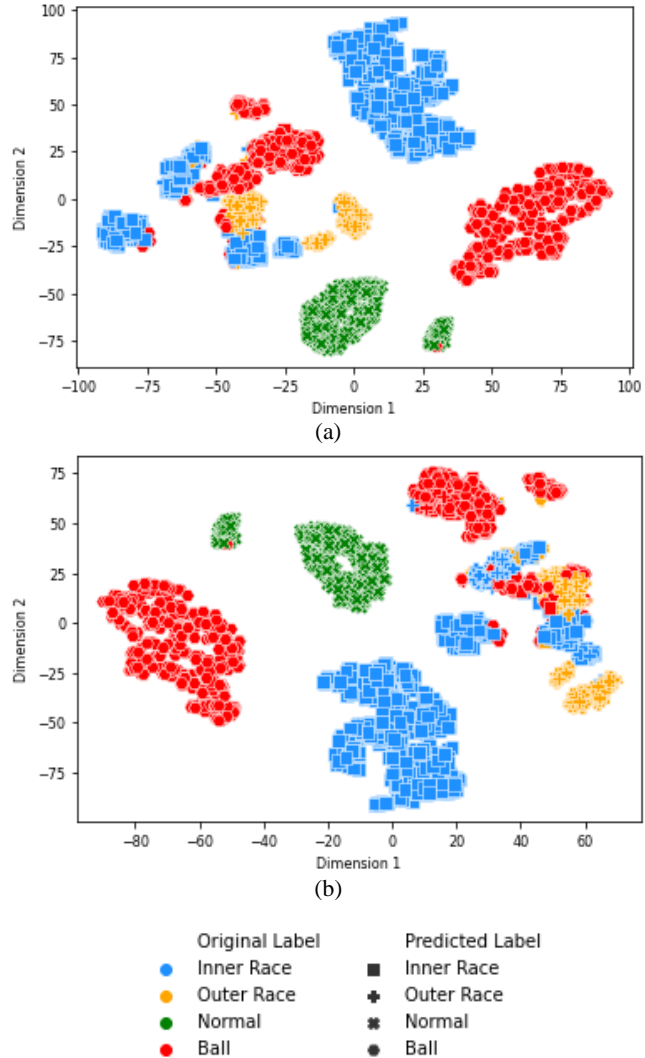

Figure 6. 2D t-SNE representation of actual classes and predicted classes by the decision tree model for a) original data points and b) crafted adversarial samples using FGSM with $\varepsilon=0.03$.

The trained models showed different performances on adversarial samples. For example, the quadratic discriminant analysis model showed an F1 score of 0.52 for $\varepsilon=0.05$, while the k-nearest neighbors obtained a score of 0.99 for the same amount of perturbation. The ensemble model's F1 scores under the different extent of perturbation suggest that using an ensemble of different models in CBM systems can prevent a dramatic decrease in the systems' performance under attack compared with the CBM systems that use just one model.

Two groups of defense strategies against adversarial machine learning have been suggested in the literature. The first group relies on modifying training algorithms, known as adversarial training, and the second group is based on the defensive mechanism to detect adversarial samples. By perturbing 20% of the training dataset using FGSM with $\varepsilon=0.03$, the models' performance improved in all ranges of perturbation (Figure 7). Therefore, in the training phase of the CBM systems taking into account adversarial samples will robust the model.

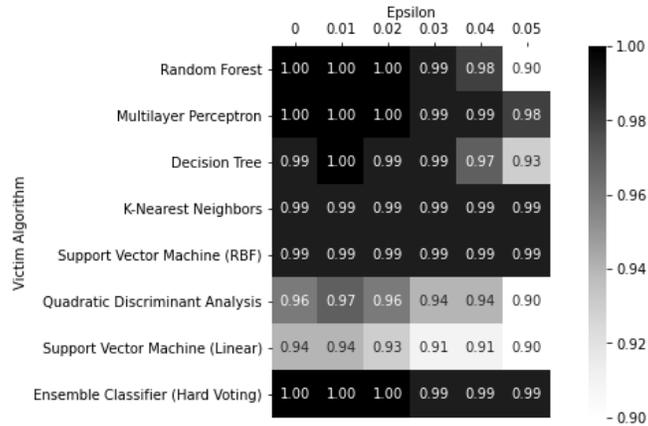

Figure 7. F1 score for different machine learning algorithms, which were trained on a dataset with 20% perturbed data points, on adversarial samples crafted using FGSM (ε: extent of the perturbation).

## 4. Conclusions

This paper investigated how adversarial machine learning techniques can be used to target condition-based maintenance (CBM) systems by deceiving machine learning models. Adversarial machine learning was previously investigated in the computer vision domain, while their application can be extended beyond that to generate adversarial attacks on CBM systems. In this study, the fast gradient sign method, which was introduced for crafting adversarial images, was used to generate adversarial input data for multiple trained supervised models that were supposed to assess the health status of a bearing.

The considered models, including random forest, multilayer perceptron, decision tree, k-nearest neighbors, non-linear support vector machine, linear support vector machine, and quadratic discriminant analysis, were able to identify the health status of the bearing with F1 scores higher than 0.95. However, all the models except k-nearest neighbors experienced a drop in their performances on the adversarial samples. The F1 score for the models under attack decreased down to 0.52 (for the quadratic discriminant analysis model) with adding perturbation with the amount of 0.05 to the original data. The observed decline in the models' performance shows their vulnerability to assess the system's health when the collected data are perturbed using adversarial machine learning.

Modifying the training procedure by perturbing 20% of the training dataset improved the models' performance on the adversarial samples by increasing the minimum F1 score to 0.9. For example, the random forest model which experienced 0.17 decline in its F1 score on the crafted adversarial samples, could make it up to a 0.1 drop after the modification. This defense strategy was considered previously in the computer vision domain. However, a comprehensive investigation on developing defense strategies for CBM systems can be a consideration for future studies.

The objectives of adversarial machine learning attacks to CBM systems are to cause unnecessary or delayed maintenance by confusing the used machine learning models. Unnecessary maintenance results in extra costs and delayed maintenance can lead to catastrophic failures. The obtained results in this paper reveal that understanding the applicability of adversarial machine learning attacks in CBM systems is necessary in order to develop more robust machine learning-based CBM systems.